# An Analysis on Automated Metrics for Evaluating Japanese-English Chat Translation

Andre Rusli, Makoto Shishido
Tokyo Denki University
{20udc91@ms, shishido@mail}.dendai.ac.jp

## 概要

  This paper analyses how traditional baseline metrics, such as BLEU and TER, and neural-based methods, such as BERTScore and COMET, score several NMT models' performance on chat translation and how these metrics perform when compared to human-annotated scores. The results show that for ranking NMT models in chat translations, all metrics seem consistent in deciding which model outperforms the others. This implies that traditional baseline metrics, which are faster and simpler to use, can still be helpful. On the other hand, when it comes to better correlation with human judgment, neural-based metrics outperform traditional metrics, with COMET achieving the highest correlation with the human-annotated score on a chat translation. However, we show that even the best metric struggles when scoring English translations from sentences with anaphoric zero-pronoun in Japanese.

## 1 Introduction

  In recent years, machine translation has grown faster to provide better and more inclusive systems for many language directions. One of the most recent projects by the research team at Facebook tries to accommodate every language so that no language is left behind. Until a few years ago, most translation systems focused only on the sentence level, unable to understand the sentence's contexts which often exist in previous sentences. Many reports have demonstrated this to be inadequate. Several studies have shown that human translators outperform NMT models when the target language translation must consider the context at the document level [1, 2]. To improve this, the research community proposed various methods and improvements to build models that could use the source- and target-side contexts [3, 4] when performing translations. By doing so, models could handle more challenging tasks which require context understanding, such as chat translations.

  For Japanese NMT, various methods have also been proposed to tackle critical challenges, especially in translating Japanese conversational texts to English. One of the most prominent challenges in discourse is the anaphoric zero-pronoun, which arises from pronouns such as subjects, objects, and possessive cases omitted from conversational sentences. Anaphoric zero-pronoun is not exclusive to Japanese but is one of the most challenging languages to resolve [5, 6]. In contrast, while the number of works of research for training better and more capable NMT models keeps increasing rapidly, progress in machine translation (MT) evaluation has been struggling to keep up. Many works of research in the MT research community still rely on traditional metrics that are outdated but widely adopted as the standard.

  Even though human judgment is still considered the best metric to measure translation quality, it is considerably more expensive and time-consuming. Therefore, automatic metrics are an indispensable part of machine translation evaluation. They provide immediate feedback during MT system development and serve as the primary metric to report the quality of MT systems. Accordingly, the reliability of metrics is critical to progress in MT research. Historically, metrics for evaluating the quality of MT have relied on assessing the similarity between an MT-generated hypothesis and a human-generated reference translation in the target language. Traditional metrics, such as BLEU [7] and TER [8], have focused on basic, lexical-level features, such as counting the number of matching n-grams between the MT hypothesis and the reference translation. On the other hand, more recently proposed metrics are primarily based on pre-trained Transformers-based language models, such as BERTScore [9] and COMET [10], to calculate the similarity of translated sentences with their target. Additionally, similarity-based metrics built using neural-based methods can help evaluate human-generated texts, such as content scoring systems [11], often found in foreign language learning settings.

This paper compares how different evaluation metrics perform on chat translation from Japanese to English. Then we analyze and emphasize the importance of using suitable metrics for measuring the performance of our translation systems depending. Additionally, we provide datasets containing the translation results of the Japanese-English Business Scene Dialogue corpus by three NMT models and a set of human-annotated scores of one of the model's translation results.

## 2 Evaluation Metrics for Japanese-English Chat Translation

In MT, many traditional baseline metrics remain popular for evaluating MT systems due to their lightweight and fast computation [10]. However, as MT systems improve over time, commonly used metrics struggle to correlate with human judgment at the segment level and fail to evaluate the highest-performing MT systems adequately, thus misleading system development with incorrect conclusions. Among these metrics, BLEU is often considered the de facto standard of MT evaluation metrics. In advanced translation, however, there are many cases in which the description of phrases differs but implies the same meaning. In these cases, BLEU will struggle to perform adequately, even if there is a system that can translate with high quality. It is noted that even though BLEU is fast and can be helpful to assist researchers and developers in quickly performing initial experiments, BLEU should not be the primary evaluation technique in NLP papers [12]. Furthermore, it is argued that, due to its limitations, the common use of BLEU over the past years has negatively affected research decisions in MT [13].

In recent years, many efforts have been conducted to propose better metrics that can measure the quality of MT systems, such as BERTScore [9] and COMET [10], which are proven to correlate better with human judgments. However, only a few provided analysis on their use for evaluating chat translation, which is notably challenging for the Japanese language with its various challenges, such as the anaphoric pronoun resolution, that occurs more frequently in spoken language than in written language [14]. Furthermore, many works that try to build models for this are evaluated using BLEU, which is often argued as insufficient for advanced models [15]. However, the performance of many recent neural-based metrics has yet to be analyzed. This paper compares and analyses results from different models measured with various metrics and how these metrics correlate with human judgment.

## 3 Experiments

In this section, we explain the data and procedure in the experiment to evaluate the quality of machine-translated sentences using various metrics and provide an analysis of how the result of each metric compares to the human evaluation score and with each other. Firstly, we use three NMT models to translate conversations from Japanese to English and store the translation results. Then, we calculate the score of each model's translation with various evaluation metrics, from traditional and commonly used metrics like BLEU to more recent neural-based metrics such as BERTScore. Furthermore, we provide an overview of a tool we use for the human evaluation of translated sentence output by the models. Using this evaluation tool, we compile human-annotated scores of machine-translated sentences from one model, measure how each metric correlates with human scores, and share our findings.

### 3.1 Dataset and NMT Model Performance

In our experiment, we use the Business Scene Dialogue (BSD) corpus [16]. We translated the whole development set of the BSD corpus (2,051 chats) using three models, namely our own Transformers-based MT model, M2M100, and MarianMT for Japanese to English language direction. Previously, only BLEU was used to evaluate the results; in this paper, we use six metrics containing both traditional baseline metrics (BLEU, TER, METEOR, and chrF) and recent neural-based metrics (BERTScore and COMET). We can see in Table 1 that even though each metric uses different approaches in evaluating the translated results, all metric seems consistent when deciding which performs the best of the three models. In other words, traditional metrics are still valid if we only want to rank which models perform better than the others, and the processing time is essential. Table 1 shows that all metrics indicate MarianMT outperforms the other two models.

**Table 1: NMT models performance measured by different metrics**

| Metric | Ours | M2M100 | MarianMT |
|---|---|---|---|
| BLEU | 0.14 | 0.12 | **0.17** |
| TER | 79.70 | 78.62 | **75.54** |
| METEOR | 0.41 | 0.39 | **0.47** |
| chrF | 34.31 | 35.07 | **39.77** |
| BERTScore | 0.92 | 0.92 | **0.93** |
| COMET | -0.022 | 0.026 | **0.225** |

### 3.2 Direct Assessment for Human Evaluation and Metrics Performance

Previous works have shown that some metrics are better than others in correlation with human judgment, which is still considered the gold truth of evaluation metrics in machine translation. In this section, we want to explore how these metrics correlate with human judgment when evaluating context-heavy chat sentences from Japanese to English in the BSD corpus. Firstly, to get the human-annotated score, we built a direct assessment tool for the manual evaluation of machine translation. The tool enables the user to view the chat dataset in a familiar chat-like user interface displaying the current segment/chat conversation along with its document-level context (previous chats), then score each translation from 0 (wrong) to 100 (perfect). It is developed following the direct assessment tool used for human evaluation on the WMT 2020 Shared Task on Chat Translation [17]. Figure 1 shows the user interface of this tool.

Using the direct assessment tool for manual evaluation, we collected human-annotated scores of 10 conversations containing 283 chats inside the development set of the BSD corpus, translated by our own MT model. We asked two human annotators to score the translation and averaged the results. Both annotators speak Japanese and English, one is a native English speaker, and the other is a native Japanese speaker. The conversations are picked arbitrarily from the development set.

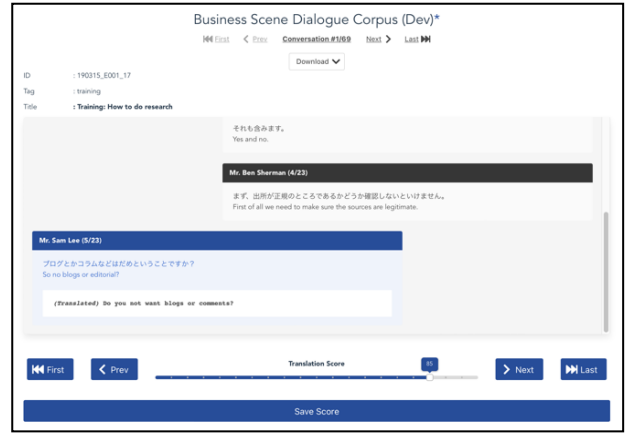

**Figure 1: User interface of the direct assessment tool**

Table 2 shows how each metric correlates with the human-annotated score for each chat translation. Neural-based metrics pre-trained with language models outperform traditional baseline metrics in correlation with human scores. It is worth noting that neural-based metrics slow down when the computation is done using the CPU only. This can be a limitation when we still experiment with architectures and hyperparameters in the initial phases of training an MT model.

**Table 2: Pearson's r on each metric**

| Metric | Pearson's r | Elapsed Time |
|---|---|---|
| SentenceBLEU | 0.2246 | **1.06 s** |
| TER | -0.1995 | 1.19 s |
| METEOR | 0.2860 | 1.23 s |
| chrF | 0.3038 | 1.07 s |
| BERTScore (F1) | 0.4342 | GPU: 6.54 s / CPU: 98.73 s |
| COMET | **0.5246** | GPU: 16.71 s / CPU: 220.71 s |

As shown in Table 2, COMET achieves the highest correlation coefficient in our experiment. Based on this, we conduct additional analysis on the results scored by COMET. The aim is to get some clues about how it evaluates translations with a specific language phenomenon, such as the anaphoric zero pronouns, which are prominent, especially in Japanese conversations/chats [18]. Table 3 provides some example sentences where COMET is supposed to give higher scores to the human-translated sentences and lower scores to machine-translated sentences in which the pronouns need to be correctly translated and could result in a completely different meaning.

Table 3: Examples where COMET gives lower score to better translations on zero-pronoun sentences

| Example #1 | | |
|---|---|---|
| Source | 最後にお話したのはいつでしたっけ？ | |
| Target | When was the last time we talked? | |
| Machine | When was the last time **you** talked? | **0.931** |
| Human | Do you remember when was the last time we talked? | 0.845 |
| Example #2 | | |
| Source | できると思います。 | |
| Target | I think so. | |
| Machine | I think **you** can. | **-0.383** |
| Human | Yes, I think I can. | -0.267 |

## 4 Conclusion

We analyze various metrics for chat translation from Japanese to English using three models on the BSD developments set. Furthermore, to determine whether the results correlate with human judgment, we compiled a dataset containing ten conversations from the BSD corpus development set (approximately 15% of the total number of exchanges) along with the translation results and their human-annotated score using Direct Assessment. By calculating the Pearson correlation coefficient (r), it can be seen that more recent metrics correlate better with human-annotated scores than traditional baseline metrics, with COMET achieving the highest correlation.

There are still, however, limitations on the current best metrics. One major challenge in translating Japanese sentences is when the model needs to decide what pronoun to use when the source sentence does not contain any. Even the best metrics often make simple mistakes in our experiments by giving high scores to translations with wrong pronouns. It is worth further research on improving the metrics to accommodate various language-specific phenomena.